\documentclass[lettersize,journal]{IEEEtran}
\usepackage{amsmath,amsfonts}
\usepackage{algorithmic}
\usepackage{algorithm}
\usepackage{array}
\usepackage[caption=false,font=normalsize,labelfont=sf,textfont=sf]{subfig}
\usepackage{textcomp}
\usepackage{stfloats}
\usepackage{url}
\usepackage{verbatim}
\usepackage{graphicx}
\usepackage{cite}
\usepackage{makecell}
\usepackage{xcolor}
\usepackage{multirow}
\usepackage{booktabs}       

\begin{document}

\title{Retinex Meets Language: A Physics-Semantics-Guided Underwater Image Enhancement Network

}

\author{Shixuan Xu, Yabo Liu, Chao Huang,~\IEEEmembership{Member,~IEEE,} Junyu Dong,~\IEEEmembership{Member,~IEEE,} and Xinghui Dong,~\IEEEmembership{Member,~IEEE}

\thanks{This study was supported in part by the National Natural Science Foundation of China (NSFC) (No. 42576200) and in part by the Key Research and Development Program of Shandong Province, China (No. 2024ZLGX06)(Corresponding author: Xinghui Dong).}
\thanks{S. Xu, Y. Liu, J. Dong and X. Dong are with the State Key Laboratory of Physical Oceanography and the Faculty of Information Science and Engineering, Ocean University of China, Qingdao, 266100. (e-mail: xushixuan@stu.ouc.edu.cn, yaboliu.ug@gmail.com, dongjunyu@ouc.edu.cn, xinghui.dong@ouc.edu.cn).}
\thanks{C. Huang is with the School of Cyber Science and Technology, Shenzhen Campus of Sun Yat-sen university, Shenzhen, China. (e-mail: huangch253@mail.sysu.edu.cn).}
}

\maketitle

\begin{abstract}

Underwater images often suffer from severe degradation caused by light absorption and scattering, leading to color distortion, low contrast and reduced visibility. Existing Underwater Image Enhancement (UIE) methods can be divided into two categories, i.e., prior-based and learning-based methods. The former rely on rigid physical assumptions that limit the adaptability, while the latter often face data scarcity and weak generalization. 
To address these issues, we propose a Physics-Semantics-Guided Underwater Image Enhancement Network (PSG-UIENet)\footnote{The data set, source code and model will be made publicly available at https://github.com/INDTLab/PSG-UIENet upon the acceptance of the paper.}, which couples the Retinex-grounded illumination correction with the language-informed guidance. This network comprises a Prior-Free Illumination Estimator and a Semantics-Guided Image Restorer. In particular, the restorer leverages the textual descriptions generated by the Contrastive Language-Image Pre-training (CLIP) model to inject high-level semantics for perceptually meaningful guidance. 
Since multi-modal UIE data sets are not publicly available, we also construct a large-scale image-text UIE data set, namely, LUIQD-TD, which contains 6,418 image-reference-text triplets. To explicitly measure and optimize semantic consistency between textual descriptions and images, we further design an Image-Text Semantic Similarity (ITSS) loss function. To our knowledge, this study makes the first effort to introduce both textual guidance and the multi-modal data set into UIE tasks. Extensive experiments on our data set and four publicly available data sets demonstrate that the proposed PSG-UIENet achieves superior or comparable performance against fifteen state-of-the-art methods.

\end{abstract}


\section{Introduction}
\label{intro}

\IEEEPARstart{U}{nderwater} images play a critical role in a diverse range of applications, including marine biology research, underwater archaeology, seabed mapping, and autonomous robot navigation \cite{tcsvt1,tcsvt2,imts1,imts2}. However, the unique optical properties of water—characterized by absorption, scattering, and suspended particles—inevitably degrade image quality. As a result, underwater images often suffer from severe artifacts, such as color distortion, low contrast, and reduced visibility. These degradations will impair the reliability of both manual analysis and automated systems, becoming a major bottleneck of underwater exploration and monitoring tasks. Therefore, Underwater Image Enhancement (UIE) has emerged as a crucial research topic with significant scientific and practical values.

\begin{figure}[t]
\centering    
\includegraphics[width=0.48\textwidth]{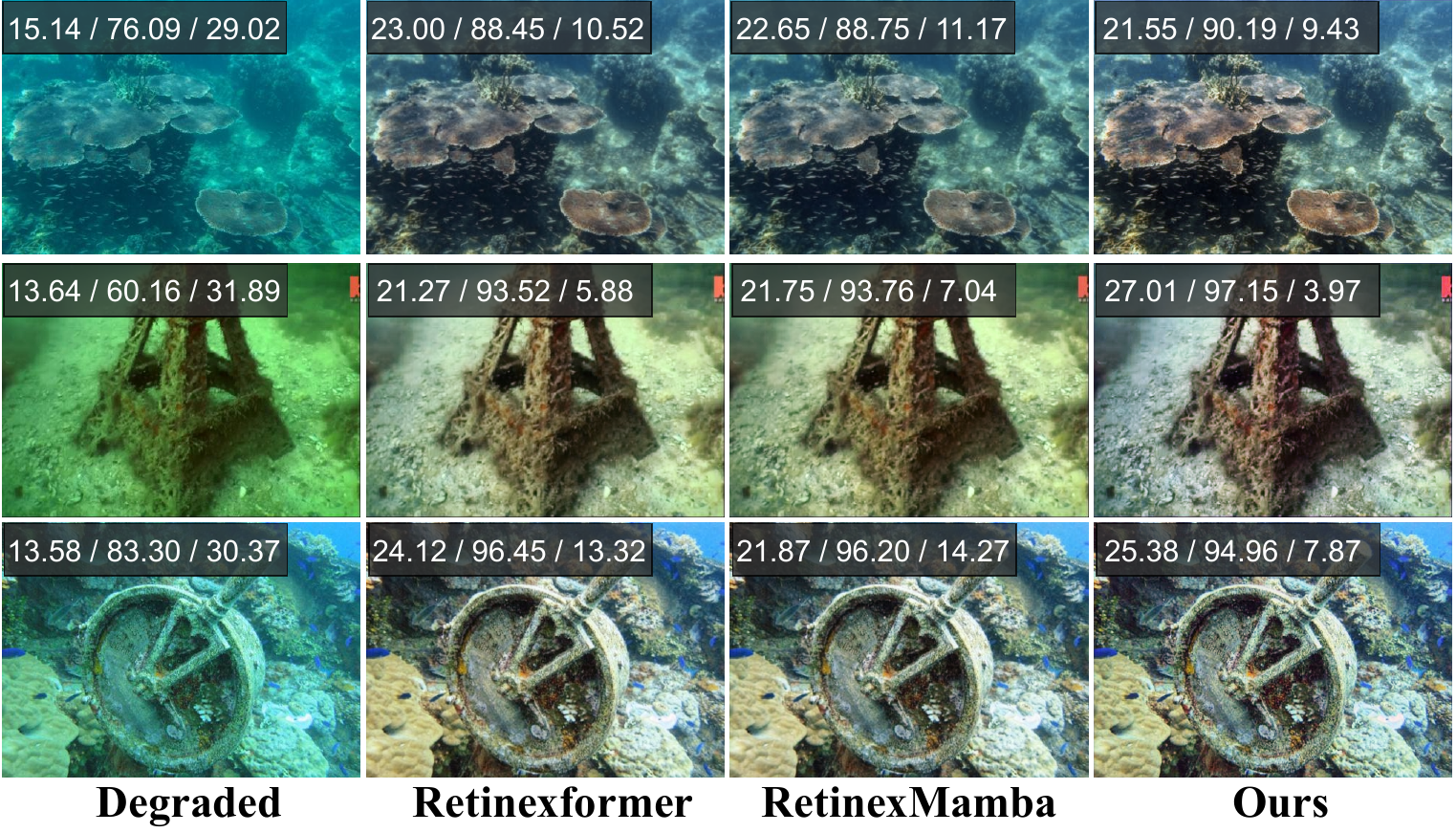}
\caption{Comparison of three Retinex-based UIE methods, including Retinexformer \cite{retinexformer}, RetinexMamba \cite{retinexmamba}, and our PSG-UIENet. In terms of each image, the Peak Signal to Noise Ratio (PSNR), Structural Similarity Index (SSIM) \cite{ssim}, and Learned Perceptual Image Patch Similarity (LPIPS) \cite{lpips} values computed between the corresponding reference image and it are shown at the top-left corner. These results highlight the effectiveness of the Retinex theory \cite{retinex_theory} in alleviating underwater image degradation.}
\label{fig:retinex}
\end{figure}


Many UIE methods have been developed over the years. Existing UIE methods can generally be categorized into two paradigms, i.e., prior-based and learning-based approaches. Prior-based methods \cite{DCP, UDCP, GDCP, water_hl, retinex, hlrp, mmle, hfm} normally rely on handcrafted physical priors or models to estimate illumination and transmission. These methods are interpretable and computationally efficient. However, they often fail to generalize across diverse underwater environments due to their reliance on strict assumptions, which may not hold under varying conditions.

On the other hand, learning-based methods \cite{waternet, ucolor, pats-uienet, puie, ushape, uranker, ccmsrnet, uwformer} generally leverage deep neural networks to directly learn complex mappings from data. For instance, Convolutional Neural Network (CNN)-based methods extract local hierarchical features to enhance degraded images, while Transformer-based approaches employ self-attention mechanisms to capture long-range dependencies. Although remarkable progress has been made, they are heavily dependent on large-scale annotated data sets. Unfortunately, existing UIE data sets normally contain limited real-world samples and lack diversity, thereby decreasing the generalization of the model trained. Thus, both paradigms face inherent limitations, underscoring the demand for more robust and adaptive UIE methods to diverse underwater environments.


In light of these limitations, recent studies have paid attention to hybrid paradigms \cite{retinexformer, ccmsrnet, pats-uienet, retinexmamba} that combine physical priors with deep learning techniques. Among these, the Retinex theory \cite{retinex_theory} has attracted significant attention due to its ability to effectively model illumination and reflectance. For example, Retinex priors were integrated into deep networks and achieved improvements in contrast, color accuracy, and structural preservation (see Fig.~\ref{fig:retinex}) \cite{retinexformer, ccmsrnet, retinexmamba}.

Furthermore, vision–language models, such as Contrastive Language-Image Pre-training (CLIP) \cite{clip}, have demonstrated exceptional potential to align visual and textual semantics, allowing high-level semantic guidance for image enhancement. These developments suggest that the joint use of physical priors and semantic information can address the limitations of traditional methods. 
However, most of the existing approaches \cite{clip-lit, hazeclip, clip-uie} often fail to fully exploit the complementary strengths of the two types of data. Specifically, the scarcity of multi-modal data sets severely limits the integration of physical priors and semantics-driven guidance in domain-specific tasks, e.g., underwater image enhancement. This limitation motivates us to explore a new framework that bridges the gap by leveraging both physical priors and text-based semantics.



To this end, we propose a novel Physics-Semantics-Guided Underwater Image Enhancement Network (PSG-UIENet). This network integrates physics-inspired priors obtained using the Retinex theory \cite{retinex_theory} with semantic guidance provided by textual descriptions, forming a unified multi-modal enhancement framework. 
Specifically, PSG-UIENet consists of a Prior-Free Illumination Estimator and a Semantics-Guided Image Restorer. Unlike existing Retinex-based methods \cite{retinexformer, retinexmamba} which rely on handcrafted priors, our illumination estimator operates without fixed assumptions. The image restorer adopts a dual-branch, mask-based encoder–decoder, enabling effective integration of visual and textual features for semantically guided enhancement.

To facilitate multi-modal learning, we also collect the first large-scale image–text UIE data set, i.e., LUIQD-TD. It is derived on top of the LUIQD \cite{pauqa} data set. In total, 6,418 image-reference-text triplets are included. Each triplet comprises a degraded image, a visually optimal reference image, and a corresponding textual description. We further propose a novel Image-Text Semantic Similarity (ITSS) loss function to explicitly enforce semantic alignment between textual descriptions and enhanced images. This loss function ensures that the enhanced image is not only visually appealing but also semantically consistent with the given textual description.



To our knowledge, this is the first study to incorporate textual descriptions into a prior-based UIE network and to build an image-reference-text data set for UIE tasks. The contributions of this study can be summarized as fourfold.
\begin{itemize}
\item We propose PSG-UIENet, the first physics-semantics-guided UIE network that integrates a prior-free illumination estimator and a semantics-driven image restorer, advancing multi-modal UIE research.
\item We construct the first multi-modal UIE data set, i.e., LUIQD-TD, which contains 6,418 image-reference-text triplets, satisfying a critical demand in multi-modal data for UIE tasks.
\item We design a dual-branch cross-modal fusion mechanism with random pixel-level masking to extract and align textual semantics with visual features. In particular, a Cross-Attention FiLM Module (CFM) is introduced for adaptive semantic fusion. 
\item We perform extensive experiments on five benchmark test sets together with fifteen state-of-the-art methods. The results not only demonstrate the superiority of our approach, but also establish new baselines for future multi-modal UIE research.
\end{itemize}

The rest of this paper is organized as follows. In Section \ref{related_work}, we review the related work. Our data set is introduced in Section \ref{dataset}. Section \ref{methods} presents our method. Experimental settings and results are reported in Sections \ref{experiment} and \ref{results}, respectively. Finally, we conclude the paper in Section \ref{conclusion}.

\section{Related Work}
\label{related_work}

\subsection{Underwater Image Enhancement Methods}
Underwater Image Enhancement (UIE) methods can be divided into prior-based and learning-based methods.  
Prior-based methods rely on physical models or handcrafted assumptions to estimate illumination or transmission, for example, Dark Channel Prior (DCP) \cite{DCP}, its underwater variants, i.e., UDCP \cite{UDCP}, and GDCP \cite{GDCP}. Although these methods are simple and interpretable, their reliance on rigid priors often limits their adaptability to complex underwater conditions and may introduce visual artifacts \cite{water_hl, water_hl_2}. To address these limitations, more general priors were proposed for UIE approaches, such as Histogram Equalization (HE) \cite{CLAHE, RGHS}, Retinex-based methods \cite{retinex, EMS_Retinex}, and multi-exposure fusion frameworks \cite{fusion1, fusion2}. Although these approaches can improve contrast and color, they usually lack scene-awareness and struggle with handling global variations in underwater degradation \cite{hlrp, mmle, hfm}.

In contrast, learning-based UIE approaches leverage deep neural networks to directly learn enhancement mappings from data. CNN-based \cite{waternet, ucolor, pats-uienet} and Transformer-based methods \cite{puie, uranker, ushape, ccmsrnet, uwformer} have demonstrated notable success by capturing local features and long-range dependencies, respectively. However, due to the scarcity of large-scale real-world underwater data sets, early models trained mainly on synthetic data suffer from domain gaps. The introduction of data sets such as UIEB \cite{waternet} and SUIM-E \cite{sguie} has improved model generalization by providing high-quality reference images for supervised training. Despite these advances, a small single-modal data set cannot adequately cover complex real-world underwater scenarios. In contrast, textual descriptions can leverage semantic information to compensate for the limitations of single-modal data, guiding the model toward producing more stable and perceptually-consistent enhancement results.


\subsection{Retinex-Based Image Enhancement Methods}
Inspired by the human visual system, Retinex theory \cite{retinex_theory} decomposes an image into reflectance and illumination components, providing strong physical interpretability for image enhancement tasks. Recently, this theory has been integrated with deep learning techniques to improve both interpretability and performance. Qi et al. \cite{ccmsrnet} proposed a two-stage framework that combines color correction and visibility enhancement with axial attention to better capture global context. Cai et al. \cite{retinexformer} introduced a Transformer-based Retinex model with illumination-guided attention, effectively reducing artifacts and enhancing contrast. In \cite{retinexmamba}, Bai et al. incorporated state-space modeling and illumination-fused attention into Retinex-inspired architectures, achieving improved efficiency and semantic consistency.

In contrast, the proposed PSG-UIENet advances both architecture and modeling. Specifically, we design a multi-scale illumination estimator that adaptively infers illumination without relying on handcrafted priors, thereby enhancing robustness across diverse underwater scenarios. In addition, we introduce high-level semantic cues encoded by textual descriptions, enabling context-aware enhancement that extends beyond conventional Retinex-based frameworks.

\subsection{Text-Guided multi-modal Image Enhancement Methods}
In recent years, multi-modal approaches that integrate vision and language have shown great potential in image enhancement and restoration \cite{clip-lit, rave, hazeclip, clip-uie}. Using the semantic understanding ability of pre-trained vision-language models, such as CLIP \cite{clip}, these approaches aim to generate enhanced images that align more closely with human perception. In \cite{clip-lit}, CLIP-based prompt learning was introduced with a ranking optimization strategy for unsupervised backlit enhancement. Chen et al. \cite{rave} used residual vectors in the embedding space of CLIP rather than handcrafted prompts, improving the usability and transferability of CLIP guidance. In \cite{hazeclip}, CLIP guidance was applied to dehazing with region-specific prompts and contrastive learning. To bridge domain gaps between synthetic and real-world underwater images, Liu et al. \cite{clip-uie} combined CLIP with diffusion models and a CLIP-classifier.

Nevertheless, existing text-guided methods typically treat textual input as a weak supervisory signal, relying heavily on fixed prompts or CLIP-based similarity loss functions. As a result, they cannot fully exploit the rich semantics of natural language and were mainly tailored for natural scene enhancement, leaving domain-specific tasks, e.g., UIE, largely unexplored. To address these issues, we propose a novel Physics-Semantics-Guided Underwater Image Enhancement Network (PSG-UIENet). This network explicitly integrates Retinex-inspired physical priors with semantic information encoded in textual descriptions, enabling fine-grained alignment between visual and textual modalities. Compared to existing prompt-based approaches, our method can perform robust, context-aware enhancement for challenging underwater scenarios.

\section{LUIQD-TD}
\label{dataset}

The availability of high-quality data sets plays a pivotal role in advancing data-driven UIE methods. However, existing data sets normally face two critical limitations that hinder the development of robust and generalizable UIE models. First, the majority of existing underwater data sets are small, which contain only a few hundred samples, restricting the practical application to training deep neural networks. Second, these data sets often lack multi-modal annotations, such as textual descriptions, further limiting the exploration of the semantics cue in UIE tasks. In this context, these limitations underscore the demand for a large-scale multi-modal data set tailored to the unique challenges in the UIE scenario.


To mitigate the scarcity of multi-modal UIE data sets, we extended the Large-Scale Underwater Image Quality Data Set (LUIQD) \cite{pauqa} by obtaining a textual description for each image. As a result, we constructed a new data set, namely, LUIQD-TD. The data set was tailored for multi-modal UIE tasks, enabling the model to jointly leverage visual content and linguistic semantics for perceptually aligned restoration and semantically guided enhancement.

\subsection{LUIQD}
The LUIQD \cite{pauqa} is a large-scale human perception inspired benchmark collected for Underwater Image Quality Assessment (UIQA). It contains 64,180 underwater images, including 6,418 original degraded real-world underwater images and 57,762 enhanced images obtained using nine traditional and deep learning-based UIE algorithms. Each image was annotated with a subjective quality score ranging from 0 to 100. These scores captured human perceptual judgments, such as color fidelity, contrast, texture clarity, visibility, and foreground recognition. Due to its scales, diversity, and reliable subjective scoring, the LUIQD can serve as a solid foundation for evaluating UIE methods, selecting reference images, and constructing downstream multi-modal tasks.

\subsection{Reference Image Selection}
For supervised UIE tasks, we constructed high-quality degraded-reference image pairs by leveraging the subjective quality scores contained in the LUIQD. To be specific, each degraded image in the LUIQD-TD was directly derived from the original image contained in the LUIQD \cite{pauqa}. In terms of each degraded image, the reference image was selected as the image that received the highest quality score among the nine enhanced images. This standardized procedure ensures an objective and reproducible reference selection. In total, we obtained 6,418 degraded-reference pairs $(\mathbf{I}^{deg}_i, \mathbf{I}^{ref}_i)$, providing reliable supervision for mapping a degraded underwater image to its perceptually optimal counterpart.

\subsection{Textual Description Annotation}
To incorporate semantic information and facilitate vision–language modeling, we annotated each degraded image with a textual description $T_i$. The description highlights the key scene elements and visual characteristics shared between the degraded and reference images, including semantic content, quality attributes, and foreground-background structure. To ensure accuracy and semantic consistency, the textual descriptions were initially generated using ChatGPT-4\footnote{\url{https://chatgpt.com/}}. We then individually reviewed and refined each description, making necessary modifications to ensure that the description faithfully captures the scene content, maintains consistent terminology, and avoids semantic ambiguity. This hybrid caption pipeline combines the scalability of automatic generation with the reliability of manual refinement, ensuring both efficiency and annotation quality. As a result, we obtained 6,418 multi-modal triplets in the form:
\begin{equation}
\mathcal{D}_i = (\mathbf{I}^{deg}_i, \mathbf{I}^{ref}_i, T_i), \quad i = 1, \ldots, 6,418.
\label{equ:dataset}
\end{equation}

\begin{figure*}[t]
\centering    
\includegraphics[width=0.95\textwidth]{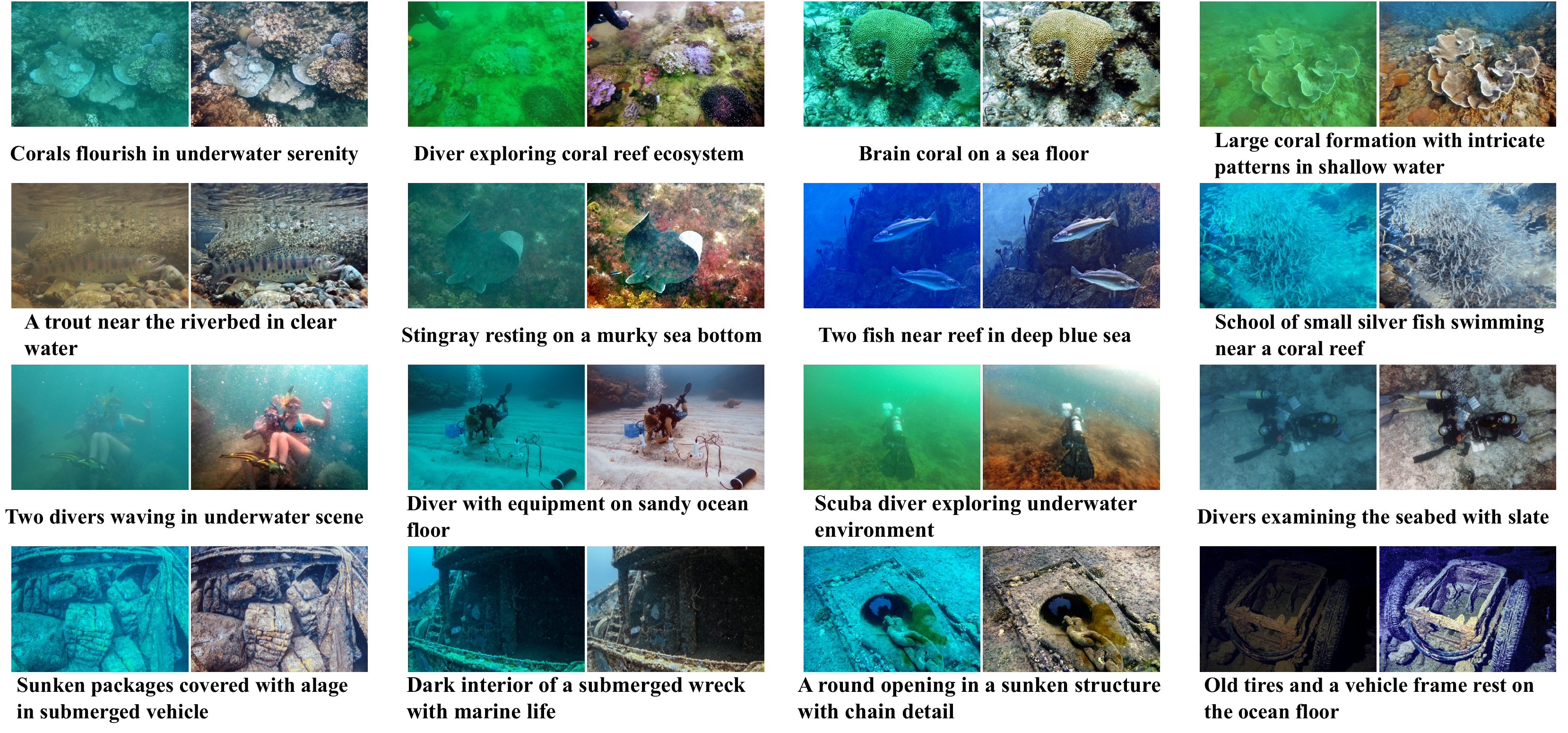}
\caption{Sixteen degraded-reference-text triplets contained in the LUIQD-TD. Each triplet consists of three components: a degraded image (top-left) and the associated reference image (top-right) and textual description (bottom). The data set spans diverse underwater scenarios, including coral reefs, marine life, divers, submerged wrecks, and underwater vehicles, thereby offering rich semantic and visual information for multi-modal UIE tasks.}
\label{fig:dataset}
\vspace{-5pt}
\end{figure*}

\begin{figure*}[t]
\centering    
\includegraphics[width=0.95\textwidth]{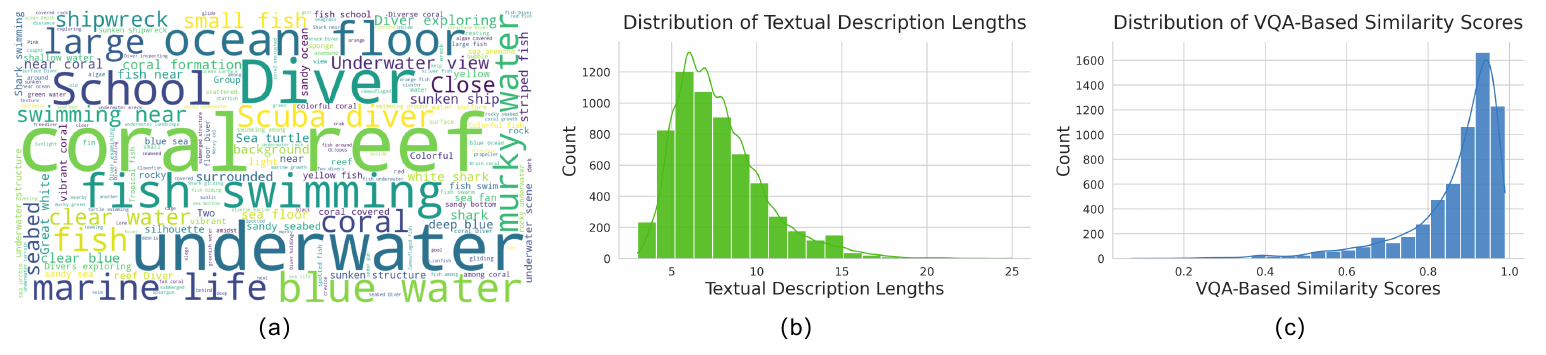}
\caption{Statistical analysis of the textual annotations in the LUIQD-TD, including (a) the distribution of word frequencies, (b) the distribution of caption lengths, and (c) the distribution of VQA-based\cite{vqa} image–text similarity scores.}
\label{fig:dataset2}
\end{figure*}

This triplet design enables text-guided UIE tasks, where a UIE model enhances a degraded image using not only visual information but also semantic cues. Sixteen triplets in the LUIQD-TD are shown in Fig.~\ref{fig:dataset}. In addition, Fig.~\ref{fig:dataset2} displays three statistical illustrations of the textual descriptions, including the distribution of word frequencies, the distribution of caption lengths and the distribution of image–text similarity scores, further demonstrating the semantic quality and consistency of the LUIQD-TD descriptions.

To our knowledge, LUIQD-TD is the first large-scale vision-language data set dedicated to UIE tasks. By providing aligned degraded-reference pairs and high-quality textual annotations, LUIQD-TD bridges the gap between low-level restoration and high-level semantics. This data set offers a novel benchmark for text-guided UIE, multi-modal supervision, and semantic evaluation, fostering the development of next-generation multi-modal UIE methods.

\section{Methodology}
\label{methods}

\begin{figure*}[t]
\centering    
\includegraphics[width=0.95\textwidth]{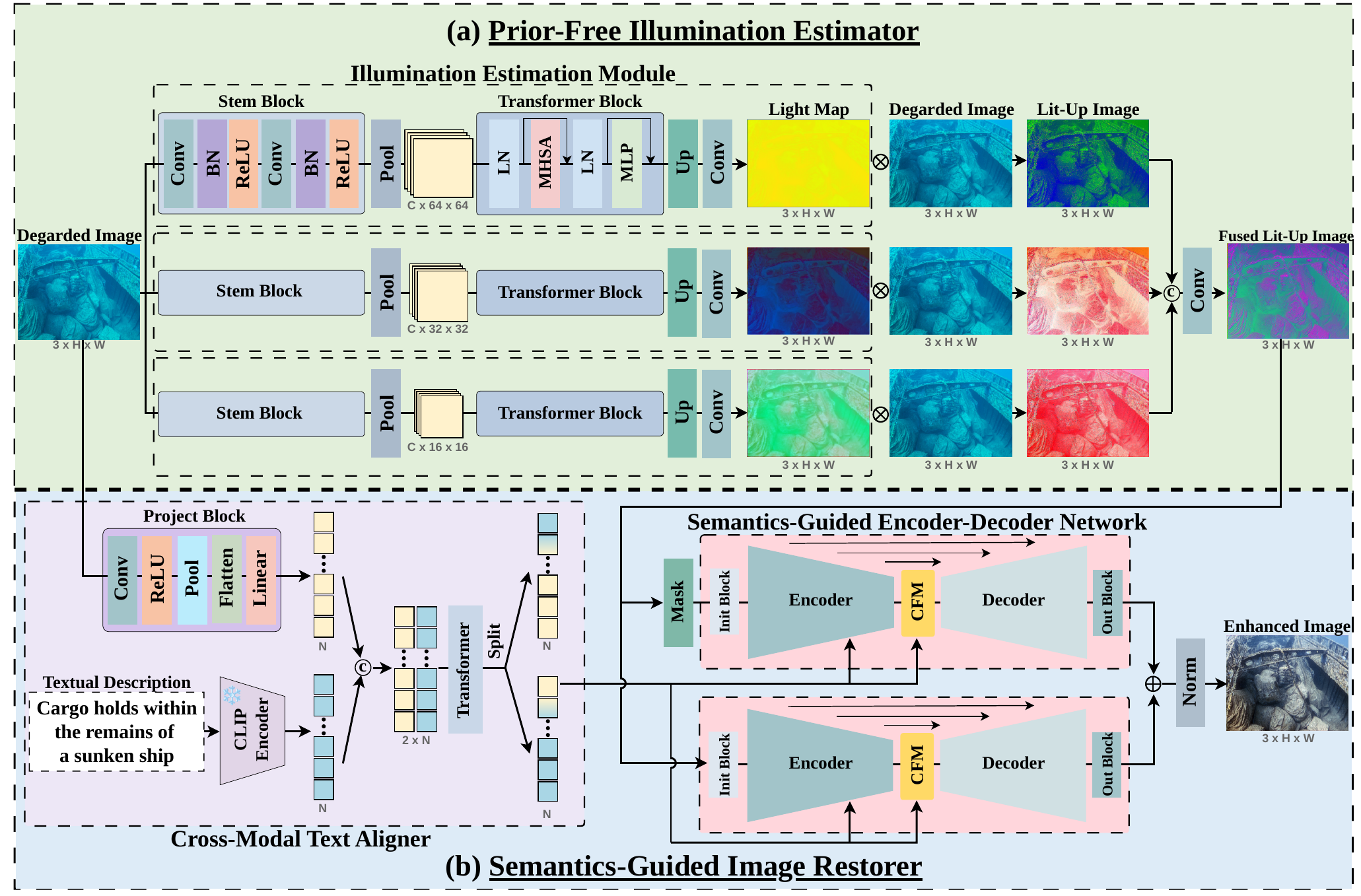}
\caption{The architecture of our PSG-UIENet, which comprises two modules: (a) a Prior-Free Illumination Estimator that generates multi-scale light-enhanced representations and (b) a Semantics-Guided Image Restorer that performs multi-modal alignment, fusion and enhancement using a dual-branch structure.}
\label{fig:all_arch}
\end{figure*}

Motivated by the Retinex theory \cite{retinex}, we propose the Physics-Semantics-Guided Underwater Image Enhancement Network (PSG-UIENet). As shown in Fig. \ref{fig:all_arch}, this network is designed to integrate physical priors with high-level semantic guidance, allowing more perceptually meaningful underwater image enhancement. PSG-UIENet comprises two main components: a prior-free illumination estimator and a semantics-guided image restorer. Compared with existing approaches\cite{pats-uienet,puie,ushape,uwformer,uranker,retinexformer,ccmsrnet,retinexmamba}, PSG-UIENet is able to utilize both the physical theory and semantic guidance, and therefore is likely to achieve superior performance in complex real-world underwater scenarios.

\subsection{Preliminaries}

The Retinex theory \cite{retinex_theory} models the Human Visual System (HVS) by decomposing an image into two components: reflectance and illumination. The former represents the intrinsic properties of the scene, while the latter indicates the environmental lighting. The decomposition can be formulated as:
\begin{equation}
\label{equ:retinex}
I(x) = R(x) \cdot L(x),
\end{equation}
where $R(x)$ denotes the reflectance, and $L(x)$ represents the illumination. The Retinex theory has inspired many image enhancement methods due to its strong physical interpretability.
%
Although its effectiveness has been demonstrated in many scenarios, the conventional formulation \cite{retinex_theory} faces significant challenges in underwater environments. The unique optical properties of water, such as nonuniform lighting, severe color cast, scattering, backscatter, and sensor noise, make the direct estimation of $R(x)$ and $L(x)$ very brittle. 

To address these issues, recent studies have extended the Retinex theory with perturbation-aware formulations \cite{retinexformer, retinexmamba} by introducing perturbation terms, to account for deviations in reflectance and illumination. Inspired by these insights, the Retinex decomposition was reformulated to better handle underwater complexities \cite{retinexformer}. Specifically, two perturbation terms, i.e, $\hat{R}$ and $\hat{L}$, were introduced to model deviations from the ideal reflectance and illumination conditions:
\begin{align}
I_{deg} &= (R + \hat{R}) \odot (L + \hat{L}) \nonumber \\
  &= R \odot L + R \odot \hat{L} + \hat{R} \odot (L + \hat{L}),
\label{equ:perturb_retinex}
\end{align}
where $\hat{R} \in \mathbb{R}^{H \times W \times 3}$ and $\hat{L} \in \mathbb{R}^{H \times W}$ represent the perturbation terms for reflectance and illumination, respectively.


To simplify the computation and mitigate the illumination imbalance, an element-wise multiplication was further applied to both sides of Eq. (\ref{equ:perturb_retinex}) with a light map $\bar{L}$ ($\bar{L} \odot L = 1$) \cite{retinexformer}:
\begin{align}
I_{deg} \odot \bar{L} &= (R \odot L + R \odot \hat{L} + \hat{R} \odot (L + \hat{L})) \odot \bar{L} \nonumber \\
                      &= R + R \odot (\hat{L} \odot \bar{L}) + (\hat{R} \odot (L + \hat{L})) \odot \bar{L}.
\label{equ:lightup}
\end{align}
As a result, the illuminated image $I_{lit}$ can be computed as:
\begin{equation}
I_{lit} = I_{deg} \odot \bar{L} = R + C,
\label{equ:simplified_lightup}
\end{equation}
where $I_{lit} \in \mathbb{R}^{3 \times H \times W}$ is the light-enhanced image, $R \in \mathbb{R}^{3 \times H \times W}$ corresponds to the ideally exposed image (i.e., the final enhanced image $I_{enh}$), and $C \in \mathbb{R}^{3 \times H \times W}$ is the residual perturbation term.




A two-stage solution was adopted on top of the reformulated decomposition for UIE tasks \cite{retinexformer}. First, $\bar{L}$ was estimated in a data-driven, prior-free manner to obtain $I_{lit}$. This step mitigated exposure and illumination imbalances, producing a well-normalized image. Second, the residual perturbation term $C$ was reduced by restoring contrast, color fidelity, and fine details. This process was guided by high-level semantic features extracted from textual descriptions. Inspired by the success that previous studies \cite{retinexformer, retinexmamba} have achieved, we also construct our network on top of the two-stage solution.


\subsection{PSG-UIENet}

To jointly leverage the physical theory and high-level semantics for perceptually meaningful UIE, we propose a Physics-Semantics-Guided Underwater Image Enhancement Network, or PSG-UIENet for short. As shown in Fig. \ref{fig:all_arch}, PSG-UIENet contains a Prior-Free Illumination Estimator and a Semantics-Guided Image Restorer. The Illumination Estimator explicitly models multi-scale illumination maps without relying on hand-crafted priors and generates an initially light-enhanced image through an adaptive fusion mechanism. To incorporate semantic information into the enhancement process, the Image Restorer leverages the semantic understanding capability of the pre-trained CLIP \cite{clip} model. The restorer consists of two parallel branches, in which one branch employs a random binary masking strategy to enhance high-level semantic understanding of visual content, while the other branch processes unmasked image features to preserve global consistency and fine details. The end-to-end image enhancement pipeline is summarized in Algorithm \ref{alg:process_ps}. By coupling the domain-agnostic physical theory with contextual semantics, PSG-UIENet simultaneously addresses low-level photometric distortions and high-level perceptual fidelity in challenging underwater scenes.

\begin{algorithm}[t]
\caption{The UIE Operation Using PSG-UIENet}
\label{alg:process_ps}
\textbf{Input}: Degraded image $I_{deg}$, textual description $T_{TD}$\\
\textbf{Output}: Enhanced image ${I}_{enh}$
\begin{algorithmic}[1]
\STATE /* Prior-Free Illumination Estimation */
\FOR{each scale $s$ in $\{16, 32, 64\}$}
    \STATE Estimate illumination maps $\bar{L}_s$ from $I_{deg}$;
    \STATE Compute lit-up image $I_{lit_{s}} = I_{deg} \odot \bar{L}_s$;
\ENDFOR
\STATE Fuse $\{I_{{lit}_{16}}, I_{{lit}_{32}}, I_{{lit}_{64}}\}$ to obtain final lit-up image $I_{lit}$;
\STATE /* Cross-Modal Text Aligner */
\STATE Flatten $I_{deg}$ into a 1D feature vector $F_{img}$;
\STATE Encode textual description $T_{TD}$ via the frozen CLIP text encoder to obtain $F_{text}$;
\STATE Concatenate $F_{img}$ and $F_{text}$ into a feature vector $F_{joint}$;
\STATE Feed $F_{joint}$ into Transformer to obtain aligned features $[F_{img}^{\prime}, F_{text}^{\prime}]$;
\STATE Extract text-aligned feature vector $F_{text}^{\prime}$ for guidance;
\STATE /* Semantics-Guided Image Restoration */
\STATE Generate mask $M_{\theta}$ from $I_{lit}$;
\STATE // \textit{Masked Branch: }
\STATE Compute $I_1 = I_{lit} \odot M_{\theta}$;
\STATE Pass $I_1$ through semantics-guided encoder-decoder network, guided by $F_{text}^{\prime}$;
\STATE Derive intermediate output $\hat{I}_1$;
\STATE // \textit{Unmasked Branch: }
\STATE Pass $I_{lit}$ through a second identical semantics-guided encoder-decoder network, guided by $F_{text}^{\prime}$;
\STATE Obtain intermediate output $\hat{I}_2$;
\STATE /* Output */
\STATE Compute fused output: $\hat{I} = \text{Norm}(\hat{I}_1 + \hat{I}_2)$;
\STATE \textbf{return} $\hat{I}$
\end{algorithmic}
\end{algorithm}

\subsubsection{Prior-Free Illumination Estimator}

As shown in Fig. \ref{fig:all_arch}(a), the Prior-Free Illumination Estimator computes multi-scale illumination maps in a data-driven manner without explicit physical priors. It uses adaptive average pooling to capture various lighting degradations at three scales, including $16 \times 16$, $32 \times 32$, and $64 \times 64$ pixels. Given a scale $s$, a single Illumination Estimation Module ($IE_s$) predicts the light map $\bar{L}_s$ according to:
\begin{equation}
\bar{L}_s = IE_s(I_{deg}), \quad s \in \{16, 32, 64\}.
\end{equation}
Each $IE_s$ contains a Stem block and a Transformer Block, which are used to encode local and global illumination characteristics, respectively.

To compute a scale-specific lit-up image, an element-wise multiplication is applied to the input image and a light map:
\begin{equation}
I_{lit_s} = I_{deg} \odot \bar{L}_s.
\end{equation}
The lit-up images obtained at all scales are then fused by a concatenation operation and a convolution operation:
\begin{equation}
I_{lit} = Conv(Concat(I_{lit_{16}}, I_{lit_{32}}, I_{lit_{64}})).
\end{equation}
The above hierarchical estimation and fusion mechanism enables our PSG-UIENet to robustly handle various lighting degradations in underwater environments and provides a well-normalized input for the subsequent semantics-guided restoration operation.

\subsubsection{Semantics-Guided Image Restorer}

To address the challenge that underwater image restoration methods normally lack the guidance of high-level textual semantics, we propose a novel Semantics-Guided Image Restorer that integrates high-level textual semantics into the restoration process. As illustrated in Fig.~\ref{fig:all_arch}(b), the restorer is built on top of a dual-branch architecture guided by textual semantics and boosted through a masking-based learning strategy. 

%



Inspired by the semantic learning mechanism of the Masked Autoencoder (MAE)~\cite{mae}, we apply a pixel-wise random binary mask $M_\theta$ with a predefined masking ratio $\theta$ to the illumination-enhanced image $I_{\text{lit}}$.
The masking process generates two images:
\begin{equation}
I_1 = I_{lit} \odot M_{\theta}, \quad I_2 = I_{lit}.
\end{equation}
$I_1$ serves as the input of the masked branch, which focuses on reconstructing occluded regions by attending to textual semantics. The masked input forces the network to rely on contextual and semantic characteristics extracted from the text. As a result, high-level understanding of the visual content is reinforced. 
In contrast, the unmasked input $I_2$ drives an unmasked branch, which operates on the entire illumination-enhanced image to preserve structural integrity and enhance fine-grained visual details.

To unify the processing of both branches, we design a Semantics-Guided Encoder-Decoder Network (SGEDN) that integrates visual and textual modalities through cross-modal fusion mechanisms. The outputs of the masked branch and the unmasked branch are denoted as $\hat{I}_1$ and $\hat{I}_2$, respectively:
\begin{equation}
\hat{I}_i = \text{SGEDN}(I_i), \quad i = 1, 2.
\end{equation}
Finally, the outputs from both branches are aggregated using an additive fusion strategy, followed by a normalization process to produce the final enhanced image:
\begin{equation}
I_{enh} = \text{Norm}(\hat{I}_1 + \hat{I}_2).
\end{equation}





The overall architecture is designed in an encoder-decoder style with a bottleneck structure. Each encoder adopts a Transformer-Conv layer for mixed global and local feature extraction, followed by a Fuse block that injects text features via a cross-modal attention mechanism. At the bottleneck layer, we introduce a Cross-Attention FiLM Module (CFM), which utilizes the global text features extracted by CLIP to generate channel-wise scaling and shifting parameters. These parameters are used to dynamically modulate the visual feature maps, further improving semantic alignment and contextual adaptability. The decoder employs residual connections to combine shallow visual details with deep fused semantics, and progressively reconstructs the image via stacked Transformer-Conv layers, yielding outputs that are both structurally faithful and semantically coherent.

\textbf{Cross-Modal Text Aligner.}  
As illustrated in Fig. \ref{fig:all_arch}(b), the Cross-Modal Text Aligner module establishes precise semantic correspondence between image features and textual representations, serving as a key component for enabling multi-modal collaborative enhancement.
While CLIP~\cite{clip} provides a generic cross-modal alignment capability, its training data includes limited underwater imagery, leading to potential semantic bias.
To mitigate this problem, we introduce a learnable projection block $PB(\cdot)$ to map raw image features into a unified semantic embedding space:
\begin{equation}
{E}_{img} = PB(I_{deg}),
\end{equation}
\begin{equation}
{E}_{text} = \phi(T_{TD}),
\end{equation}
where $\phi(\cdot)$ represents the frozen text encoder of CLIP, producing global text embeddings from the input description $T_{TD}$. 

To enable deep interaction between these cross-modal embeddings, we employ a Transformer encoder equipped with multi-head attention mechanisms. Specifically, the image and text embeddings are concatenated and processed as follows:
\begin{align}
({E}^{'}_{img},F^{'}_{text}) &= CMTA({E}_{img},{E}_{text}) \nonumber\\
&=T_{MHA}(Concat({E}_{img},{E}_{text})),
\label{equ:CMTA}
\end{align}
where $CMTA(\cdot)$ denotes the cross-modal text aligner module, and $T_{MHA}(\cdot)$ refers to the Transformer encoder equipped with multi-head attention mechanisms. As a critical bridging component in the proposed PSG-UIENet, this module provides a precise alignment foundation for subsequent semantics-guided enhancement, thereby improving the semantic consistency and perceptual quality of the enhanced results.



\begin{figure*}[h]
\centering    
\includegraphics[width=0.95\textwidth]{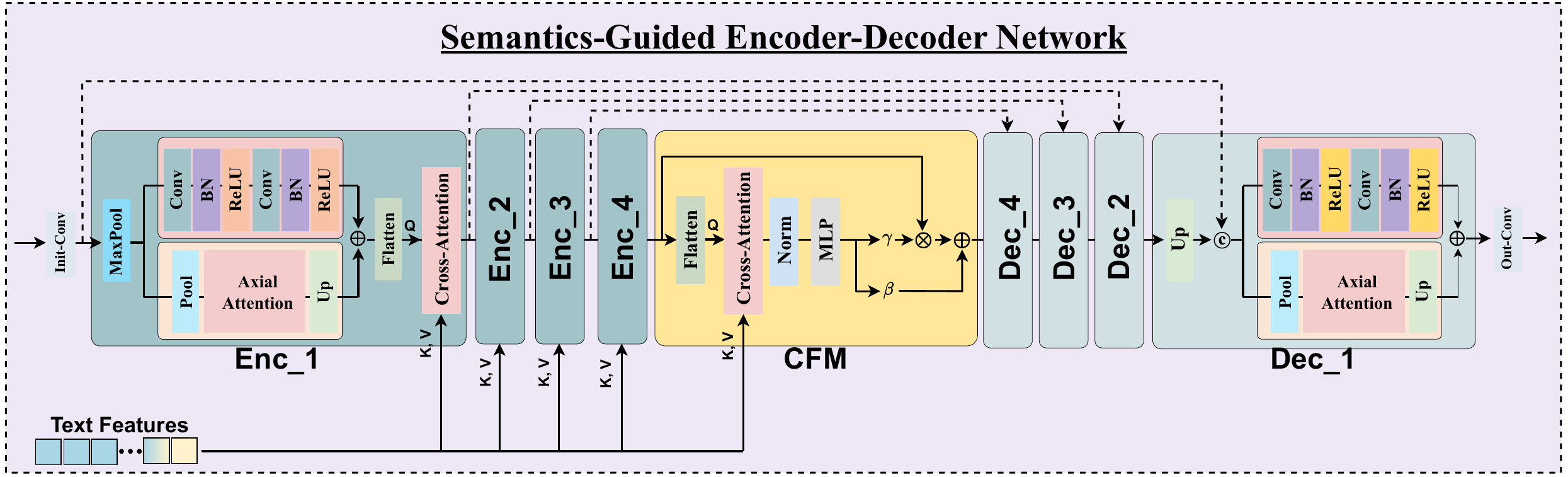}
\caption{The architecture of the Semantics-Guided Encoder-Decoder Network. This network is built on top of a symmetric encoder-decoder network, which consists of a series of Transformer-Conv layers for joint global-local feature extraction. In addition, the cross-modal attention mechanism and a Cross-Attention FiLM Module (CFM) integrate textual semantics, enabling progressive image reconstruction with semantic and visual fusion.}
\label{fig:ed_arch}
\end{figure*}

\textbf{Semantics-Guided Encoder-Decoder Network.} 
As illustrated in Fig.~\ref{fig:ed_arch}, the Semantics-Guided Encoder-Decoder Network (SGEDN) is designed to integrate visual and textual modalities for underwater image restoration. It follows an encoder-bottleneck-decoder architecture, where the encoder extracts hierarchical visual features, the bottleneck dynamically fuses visual and textual information, and the decoder progressively reconstructs the enhanced image. The core innovation lies in the bottleneck, where we introduce a \textit{Cross-Attention FiLM Module (CFM)} to achieve fine-grained semantic integration and adaptive feature modulation.

The encoder processes the input image through a sequence of Transformer-Conv layers, which are designed to capture both global and local features:
\begin{align}
F^{'}_{img} &= TC(F_{img}) \nonumber \\
&=T_{AST}(F_{img}) + Conv_{D}(F_{img}),
\end{align}
where ${Conv}_D(\cdot)$ consists of two consecutive $3 \times 3$ convolutional layers, each of which is followed by Batch Normalization and ReLU activation, and $T_{AST}(\cdot)$ denotes the axial self-attention mechanism~\cite{ccmsrnet}. This combination ensures effective extraction of both spatial details and global contextual information, which are essential for handling the complex degradations in underwater images.

The encoded image features $F^{'}_{img}$ are then fused with the aligned text features $F^{'}_{text}$, obtained from the Cross-Modal Text Aligner, through a cross-modal fuse module based on multi-head cross-attention:
\begin{align}
F_{fuse} &= Fuse(F^{'}_{img}, F^{'}_{text}) \nonumber\\
&=LayerNorm({CA}(F^{'}_{img}, F^{'}_{text}) + F^{'}_{img}),
\end{align}
where $CA(\cdot)$ denotes the multi-head cross-attention operation, which leverages image features as queries and text features as keys and values to inject semantic information into the visual representation. The fused features are passed through a residual connection followed by layer normalization to preserve original visual context while injecting semantic information.

The decoder mirrors the encoder using stacked Transformer-Conv layers and uses residual connections to merge shallow visual details with deep semantic features. This progressive reconstruction process yields outputs with both structural fidelity and semantic coherence, enabling high-quality, perceptually aligned enhancement results.

\textbf{Cross-Attention FiLM Module.} To facilitate more effective cross-modal semantic integration at the bottleneck of SGEDN during image restoration, we propose a Cross-Attention FiLM Module (CFM) as an extension to the conventional Feature-wise Linear Modulation (FiLM) \cite{film} mechanism.
Although this mechanism generates channel-wise scaling ($\gamma$) and shifting ($\beta$) parameters directly from class labels or static textual embeddings, it lacks the capacity to explicitly model the interaction between visual content and corresponding textual semantics, thereby limiting its applicability in multi-modal scenarios.

To address this limitation, the CFM augments FiLM with a cross-attention stage that aligns image features with global textual cues before producing modulation parameters. Specifically, the image features act as queries, while text features are used as keys or values in a multi-head cross-attention block.
The resulting features encode the relevance between visual regions and semantic tokens, which are then aggregated via Global Average Pooling (GAP) and passed through a Multi-layer Perceptron (MLP) to generate modulation parameters:
\begin{equation}
(\gamma, \beta) = {MLP}({GAP}({CA}(F_{{img}}, F^{'}_{{text}}))).
\end{equation}

The $\gamma$ and $\beta$ vectors are reshaped and applied to the original image features through channel-wise affine transformation:
\begin{equation}
F^{'}_{fuse} = F_{img} \odot \gamma + \beta.
\end{equation}
where $\odot$ denotes element-wise multiplication along the channel dimension. In contrast to the original FiLM \cite{film} mechanism, which statically conditions the modulation on fixed embeddings, the proposed CFM dynamically infers modulation parameters from rich cross-modal interactions. As a result, semantic information can be injected into the restoration process in a more adaptive and finer-grained manner, ultimately improving both the semantic coherence and perceptual fidelity of the enhanced underwater images.

\subsection{Loss Function}

To perform high-quality underwater image restoration, we design a composite loss function, including a Mean Squared Error (MSE) loss, an SSIM \cite{ssim} loss, a perceptual loss and an Image-Text Semantic Similarity (ITSS) loss, which balance pixel-level fidelity, structural consistency, perceptual realism and semantic alignment, respectively.

\subsubsection{ITSS Loss}
To explicitly incorporate high-level semantic guidance, we propose a novel ITSS loss function, $\mathcal{L}_{ITSS}$. Unlike traditional loss functions that rely solely on visual references, this function ensures that the enhanced image semantically aligns with a corresponding textual description $T$, while maintaining consistency with the degraded image $I_{\text{deg}}$. This is particularly critical for underwater scenes, where semantic cues are often degraded or obscured.
We leverage a pre-trained CLIP model~\cite{clip} to encode image and text embeddings into a shared latent space. Given that $\phi(\cdot)$ denote the CLIP encoder, and $\cos(\cdot, \cdot)$ represent the cosine similarity function, the ITSS loss can be formulated as:
\begin{equation}
\mathcal{L}_{{ITSS}} = \left| \cos(\phi(I_{{enh}}), \phi(T)) - \cos(\phi(I_{{ref}}), \phi(T)) \right|,
\end{equation}

This formulation introduces cross-modal supervision and serves as a semantic anchor that aligns the enhancement process with textual semantics. It is particularly beneficial for underwater scenes, where semantic cues are often lost due to severe degradation. By embedding high-level textual guidance into the optimization process, $\mathcal{L}_{{ITSS}}$ enables the model to produce enhanced images that are both visually plausible and semantically meaningful.

\subsubsection{Total Loss} 
The loss function used to train the proposed SG-UIENet is a weighted combination of the four components:
\begin{equation}
\mathcal{L}_{{total}} = \mathcal{L}_{{MSE}} + \mathcal{L}_{{SSIM}} + \alpha \cdot \mathcal{L}_{{Perceptual}} + \beta \cdot \mathcal{L}_{{ITSS}},
\end{equation}
where $\alpha$ and $\beta$ are hyperparameters used to balance the contributions of the perceptual and semantic alignment terms. Here, $\mathcal{L}_{{MSE}}$ ensures pixel-wise accuracy, while $\mathcal{L}_{{SSIM}}$ emphasizes structural consistency.  $\mathcal{L}_{{Perceptual}}$ compares high-level features extracted from the pre-trained VGG network\cite{vggnet}, which further improves perceptual quality. Finally, $\mathcal{L}_{{ITSS}}$ is specifically designed to enforce image-text semantic consistency, serving as a unique component tailored for multi-modal underwater image enhancement.

\section{Experimental Settings}
\label{experiment}

In this section, we will briefly introduce the baselines, data sets, evaluation metrics and implementation details utilized in our experiments.

\subsection{Baselines}
We compared the proposed method with six prior-based approaches, including UDCP \cite{UDCP}, Retinex \cite{retinex}, HE \cite{he}, HLRP \cite{hlrp}, MMLE \cite{mmle} and HFM \cite{hfm}.  We also compared our method with four learning-based approaches, which did not use the Retinex theory \cite{retinex}, including PUIE-Net \cite{puie}, U-Transformer \cite{ushape}, UWFormer \cite{uwformer} and PATS-UIENet \cite{pats-uienet}, three learning-based approaches which were developed on top of the Retinex theory \cite{retinex}, including Retinexformer \cite{retinexformer}, CCMSRNet \cite{ccmsrnet} and RetinexMamba\cite{retinexmamba}, and two CLIP-based approaches, including CLIP-LIT \cite{clip-lit} and CLIP-UIE \cite{clip-uie}.


\subsection{Data Sets}
We conducted a series of experiments on four publicly available real-world underwater image data sets, including \textit{LUIQD-TD}, \textit{UIEB} \cite{waternet}, \textit{SUIM-E} \cite{sguie}, and \textit{SQUID} \cite{water_hl}. The \textit{LUIQD-TD} contains a total of 6,418 image pairs. Since 194 reference images were selected as the associated degraded images according to human perceptual scores, these pairs were removed, resulting in a set of 6,224 image pairs. The UIEB data set originally included 890 image pairs, with each reference image manually selected from the results of 12 UIE methods. Similarly, ten pairs were removed, remaining 880 pairs in total. The SUIM-E \cite{sguie} data set consists of 1,525 degraded image pairs and a test set of 110 labeled images. The SQUID \cite{water_hl} data set, captured using a stereo camera, includes 114 underwater images.

We randomly split the LUIQD-TD into training, validation and testing sets at a ratio of 8:1:1. The testing set, consisting of 622 image pairs, referred to as Test-L622. In the UIEB \cite{waternet} data set, 80 image pairs were randomly selected as a second test set, named Test-U80. An additional 60 challenging images without reference images were included as the third test set, namely, Test-C60. The fourth test set, Test-S110, consists of 110 labeled image pairs from the SUIM-E \cite{sguie} data set. Finally, 53 images captured by the right-hand camera from the SQUID \cite{water_hl} data set were randomly selected and used as the fifth test set, referred to as Test-R53.

\subsection{Evaluation Metrics}
For the test data with reference images, we used PSNR, SSIM \cite{ssim}, and LPIPS \cite{lpips} to evaluate the quality difference between an enhanced image and the corresponding reference image. Regarding the test data without reference images, we utilized the Perception-Aware Underwater Image Quality Assessment (PAUQA) \cite{pauqa} and the Underwater Image Fidelity (UIF) \cite{uif} as the primary no-reference quality assessment metrics, because some studies \cite{sguie, uranker, pauqa} have reported that traditional metrics such as Underwater Color Image Quality Evaluation (UCIQE) \cite{uciqe} and Underwater Image Quality Measure (UIQM) \cite{uiqm} normally exhibit significant inconsistencies with human visual perception.

\subsection{Implementation Details}
We implemented PSG-UIENet using PyTorch and conducted all experiments on an NVIDIA RTX 3090 GPU with the Ubuntu 20.04 operating system. During the training stage, all degraded and reference images were resized to a resolution of 256$\times$256 pixels. We trained PSG-UIENet on the training set of our LUIQD-TD. The training operation was conducted using Distributed Data Parallelism (DDP) to accelerate convergence and Automatic Mixed Precision (AMP) was also enabled for computational efficiency. The network was optimized using the AdamW optimizer, with an initial learning rate of $1 \times 10^{-4}$ and a batch size of 4. The total number of training epochs was set to 100. 
The CLIP~\cite{clip} model used in our method is ViT-B/32, with frozen weights during the training process. With regard to the semantic-guided image restorer, we adopted a pixel-wise random masking strategy with a fixed masking ratio of $\theta=0.5$. The two hyperparameters of the loss function, i.e., $\alpha$ and $\beta$, were empirically set to 0.1 and 0.0001, respectively. Since test images lack textual descriptions, we used a default prompt “An underwater image” as the input of the text encoder of CLIP, to maintain consistency across the testing pipeline.

\section{Experimental Results}
\label{results}

In this section, we will report the results obtained in the UIE experiments and the ablation study.

\subsection{Full-Reference Quantitative Evaluation}
Since Test-L622, Test-U80 and Test-S110 contain reference images, we conducted a full-reference quantitative evaluation experiment on the proposed method and 15 baselines using PSNR, SSIM \cite{ssim} and LPIPS \cite{lpips}. The results are summarized in Table \ref{tab:ref}. As can be seen, our PSG-UIENet normally achieved the best performance across the three metrics on each test set. Compared with four Retinex-based methods (including Retinex \cite{retinex}, Retinexformer \cite{retinexformer}, CCMSRNet \cite{ccmsrnet}, and RetinexMamba \cite{retinexmamba}) and two CLIP-based approaches (i.e., CLIP-LIT \cite{clip-lit} and CLIP-UIE \cite{clip-uie}), our method also demonstrated advantages.

\begin{table*}[htb]
\begin{center}
\caption{The results of the full-reference quantitative evaluation of the proposed PSG-UIENet and 15 baselines on three real-world test sets. The best and second best results are highlighted in the \textcolor{red}{\textbf{Red Bold}} and \textcolor{blue}{\textit{Blue Italic}} fonts, respectively.}
\label{tab:ref}
\setlength{\tabcolsep}{3.8mm}{
\begin{tabular}{c|ccc|ccc|ccc}
    \toprule
    \multirow{2}{*}{Method} & \multicolumn{3}{c|}{Test-T622} & \multicolumn{3}{c|}{Test-U80} & \multicolumn{3}{c}{Test-S110}  \\
    \cmidrule{2-10}
      & PSNR$\uparrow$ &SSIM$\uparrow$ & LPIPS$\downarrow$ & PSNR$\uparrow$ & SSIM$\uparrow$ & LPIPS$\downarrow$ & PSNR$\uparrow$&SSIM$\uparrow$ & LPIPS$\downarrow$  \\
      \midrule
    UDCP \cite{UDCP}    & 11.51 & 47.60 & 26.94 & 9.51 & 33.66 & 41.74 & 10.07 & 34.29 & 37.66 \\
    Retinex \cite{retinex}     & 10.56 & 58.10 & 39.12 & 10.50 & 61.31 & 36.49 & 8.32 & 53.46 & 40.41 \\
    HE \cite{he}     & 16.01 & 71.30 & 28.07 & 16.60 & 77.86 & 25.67 & 15.41 & 74.79 & 30.05 \\
    HLRP \cite{hlrp}    & 13.00 & 23.83 & 33.11 & 13.56 & 22.76 & 33.90 & 12.55 & 29.24 & 33.18 \\
    MMLE \cite{mmle}     & 17.07 & 72.11 & 25.41 & 18.56 & 76.21 & 22.57 & 17.32 & 77.39 & 22.77 \\
    HFM \cite{hfm}    & 16.92 & 77.73 & 25.18 & 18.35 & 83.01 & 19.70 & 15.62 & 77.28 & 26.06 \\
    \midrule
    PUIE-Net \cite{puie}  & 23.01 & 89.03 & 10.20 & 20.45 & 87.23 & 15.29 & 21.71 & 89.89 & 9.47 \\
    U-Transformer \cite{ushape}  & 19.97 & 57.70 & 44.82 & 20.82 & 72.42 & 30.20 & 20.34 & 68.74 & 31.07 \\
    UWFormer \cite{uwformer}      & 22.65 & 87.69 & 9.92 & 19.17 & 83.82 & 17.79 & \textcolor{blue}{\textit{22.69}} & \textcolor{blue}{\textit{91.48}} & \textcolor{blue}{\textit{8.29}}\\
    PATS-UIENet \cite{pats-uienet}     & 22.95 & 88.42 & 10.39 & 21.64 & \textcolor{blue}{\textit{88.92}} & \textcolor{blue}{\textit{12.62}} & 22.68 & 91.03 & 8.58\\
    \midrule
    Retinexformer \cite{retinexformer}  & 23.24 & 90.06 & \textcolor{red}{\textbf{9.02}} & 20.71 & 87.83 & 14.19 & 22.27 & 90.99 & 8.67 \\
    CCMSRNet \cite{ccmsrnet}      & 23.10 & 88.55 & 11.31 & \textcolor{blue}{\textit{21.83}} & 88.42 & 13.62 & 22.36 & 90.14 & 9.63 \\
    RetinexMamba \cite{retinexmamba}    & \textcolor{blue}{\textit{23.42}} & \textcolor{blue}{\textit{90.09}} & \textcolor{blue}{\textit{9.08}} & 20.20 & 87.36 & 14.97 & 22.47 & 91.26 & 8.55 \\
    \midrule    
    CLIP-LIT \cite{clip-lit}    & 13.39 & 70.51 & 25.04 & 11.17 & 64.42 & 42.00 & 13.65 & 76.71 & 23.10 \\
    CLIP-UIE \cite{clip-uie}    & 18.63 & 67.05 & 35.97 & 18.51 & 74.86 & 30.94 & 18.66 & 75.55 & 26.63 \\
    \midrule
    PSG-UIENet (Ours)  & \textcolor{red}{\textbf{24.07}} & \textcolor{red}{\textbf{90.19}} & 9.11 & \textcolor{red}{\textbf{23.01}} & \textcolor{red}{\textbf{90.76}} & \textcolor{red}{\textbf{10.49}} & \textcolor{red}{\textbf{23.60}} & \textcolor{red}{\textbf{92.37}} & \textcolor{red}{\textbf{7.41}} \\    
    \bottomrule
\end{tabular}}
\end{center}
\end{table*}

\subsection{Non-reference Quantitative Evaluation}
We further used two non-reference image quality assessment metrics, PAUQA \cite{pauqa} and UIF \cite{uif}, to evaluate the performance of our method and 15 baselines on five test sets, including Test-L622, Test-U80, Test-S110, Test-C60 and Test-R53. The results are presented in Table~\ref{tab:nonref}. It can be seen that our method generally achieved comparable results with the best baseline across those test sets. Specifically, CCMSRNet \cite{ccmsrnet} produced the highest score on the five test sets, while our method consistently ranked second across these data sets, in terms of the PAUQA metric. Regarding the UIF metric, CLIP-LIT \cite{clip-lit} achieved the best performance across the five test sets. Although our method did not always secure the second-best position with regard to this metric, it still demonstrated competitive performance.

\begin{table*}[htb]
\begin{center}
\caption{The results of the non-reference quantitative evaluation of the proposed PSG-UIENet and 15 baselines on five real-world test sets. The best and second best results are highlighted in the \textcolor{red}{\textbf{Red Bold}} and \textcolor{blue}{\textit{Blue Italic}} fonts, respectively.}
\label{tab:nonref}
\setlength{\tabcolsep}{3mm}{
\begin{tabular}{c|cc|cc|cc|cc|cc}
    \toprule
    \multirow{2}{*}{Method} & \multicolumn{2}{c|}{Test-L622} & \multicolumn{2}{c|}{Test-U80} & \multicolumn{2}{c|}{Test-S110} & \multicolumn{2}{c|}{Test-C60}  & \multicolumn{2}{c}{Test-R53}                \\
    \cmidrule{2-11}
      & PAUQA$\uparrow$ & UIF$\uparrow$ & PAUQA$\uparrow$ & UIF$\uparrow$ & PAUQA$\uparrow$ & UIF$\uparrow$ & PAUQA$\uparrow$ & UIF$\uparrow$ & PAUQA$\uparrow$ & UIF$\uparrow$ \\
      
    \midrule
    UDCP \cite{UDCP}    & 36.71 & 55.91 & 29.73 & 36.40 & 29.92 & 32.01 & 28.19 & 25.41 & 33.07 & 3.24 \\
    Retinex \cite{retinex}     & 36.83 & 46.41 & 34.38 & 49.83 & 31.79 & 39.43 & 29.20 & 49.85 & 37.52 & 2.61 \\
    HE \cite{he}     & 41.32 & 45.10 & 37.82 & 44.04 & 34.26 & 45.01 & 34.69 & 41.50 & 34.95 & 1.67 \\
    HLRP \cite{hlrp}    & 36.83 & 0.57 & 38.12 & 0.59 & 37.33 & 0.91 & 30.16 & 3.57 & 28.53 & 1.26 \\
    MMLE \cite{mmle}     & 45.26 & 37.88 & 43.49 & 30.99 & 43.15 & 35.05 & 37.06 & 24.95 & 33.15 & 24.67 \\
    HFM \cite{hfm}    & 42.78 & 50.13 & 42.24 & 55.31 & 37.55 & 44.46 & 36.76 & 49.80 & 32.45 & 31.50 \\
    
    \midrule
    PUIE-Net \cite{puie}  & 49.78 & 62.60 & 45.42 & \textcolor{blue}{\textit{63.91}} & 49.15 & 59.80 & 41.45 & \textcolor{blue}{\textit{52.76}}& 41.23 & \textcolor{blue}{\textit{43.56}} \\
    U-Transformer \cite{ushape}  & 45.82 & 24.56 & 45.26 & 39.43 & 45.97 & 25.33 & 41.47 & 30.52 & 40.51 & 19.25 \\
    UWFormer \cite{uwformer}      & 48.44 & 53.23 & 42.41 & 47.98 & 48.64 & 56.10 & 38.70 & 37.39 & 40.31 & 36.27 \\
    PATS-UIENet \cite{pats-uienet}     & 49.99 & 67.66 & 45.58 & 58.63 & 48.64 & 62.53 & 41.11 & 47.77 & 39.02 & 43.51 \\
    
    \midrule
    Retinexformer \cite{retinexformer}  & 51.02 & 66.94 & 45.43 & 61.63 & 49.34 & 60.05 & 41.45 & 60.49 & 40.90 & 37.70 \\
    CCMSRNet \cite{ccmsrnet}     & \textcolor{red}{\textbf{52.04}} & 60.91 & \textcolor{red}{\textbf{47.66}} & 53.51 & \textcolor{red}{\textbf{51.74}} & 55.83 & \textcolor{red}{\textbf{42.86}} & 43.79 & \textcolor{red}{\textbf{42.25}} & 34.39 \\
    RetinexMamba \cite{retinexmamba}    & 50.91 & 66.20 & 45.09 & 61.44 & 49.03 & 58.40 & 41.34 & 39.47 & 40.48 & 35.42 \\
    
    \midrule    
    CLIP-LIT \cite{clip-lit}    & 46.76 & \textcolor{red}{\textbf{70.89}} & 37.92 & \textcolor{red}{\textbf{70.93}} & 45.71 & \textcolor{red}{\textbf{66.64}} & 37.44 & \textcolor{red}{\textbf{54.65}} & 36.81 & \textcolor{red}{\textbf{57.50}} \\
    CLIP-UIE \cite{clip-uie}    & 45.79 & 61.00 & 41.33 & 54.87 & 44.72 & 53.98 & 37.93 & 41.12 & 36.94 & 30.24 \\
    
    \midrule
    PSG-UIENet (Ours)  & \textcolor{blue}{\textit{51.34}} & \textcolor{blue}{\textit{67.70}} & \textcolor{blue}{\textit{46.32}} & 59.46 & \textcolor{blue}{\textit{49.37}} & \textcolor{blue}{\textit{62.60}} & \textcolor{blue}{\textit{41.63}} & 49.87 & \textcolor{blue}{\textit{41.32}} & 42.65 \\
    \bottomrule
  \end{tabular}}
\end{center}
\end{table*}

\subsection{Qualitative Analysis} 

\begin{figure*}[t]
\centering    \includegraphics[width=0.95\textwidth]{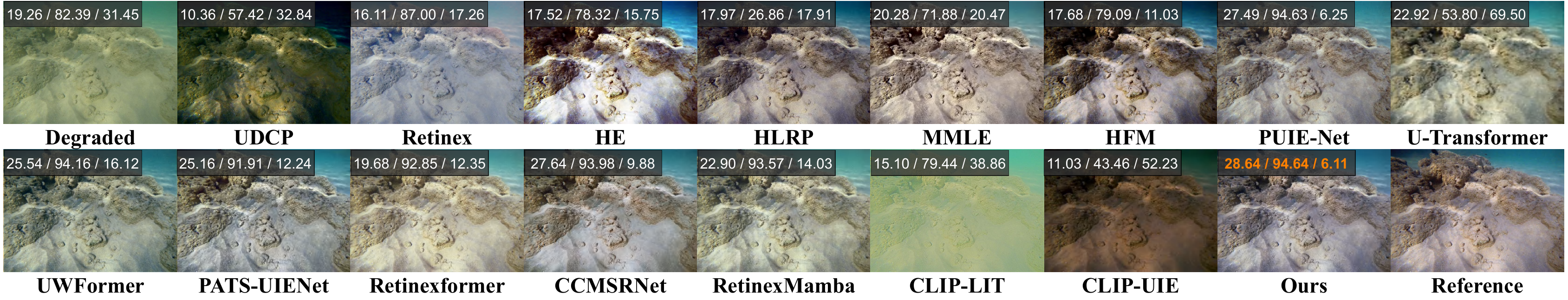}
 \caption{The results produced by 15 baselines and our method in terms of a degraded image in the Test-L622 test set. Here, the PSNR, SSIM and LPIPS values, computed between a degraded or enhanced image and the reference image, are shown at the top-left corner of the image.}
\label{fig:results_l622}
\end{figure*}

\begin{figure*}[!t]
\centering    \includegraphics[width=0.95\textwidth]{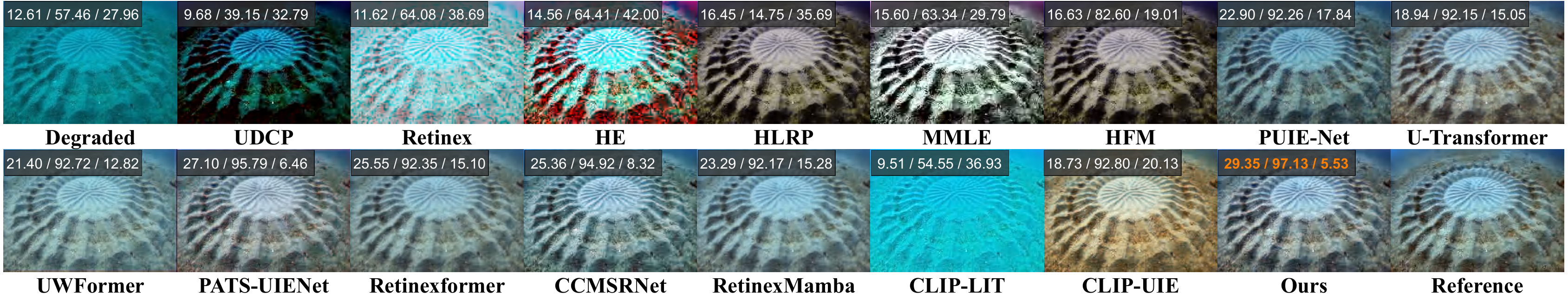}
\caption{The results produced by 15 baselines and our method in terms of a degraded image in the Test-U80 test set. Here, the PSNR, SSIM and LPIPS values, computed between a degraded or enhanced image and the reference image, are shown at the top-left corner of the image.}
\label{fig:results_u80}
\end{figure*}

\begin{figure*}[!t]
\centering    \includegraphics[width=0.95\textwidth]{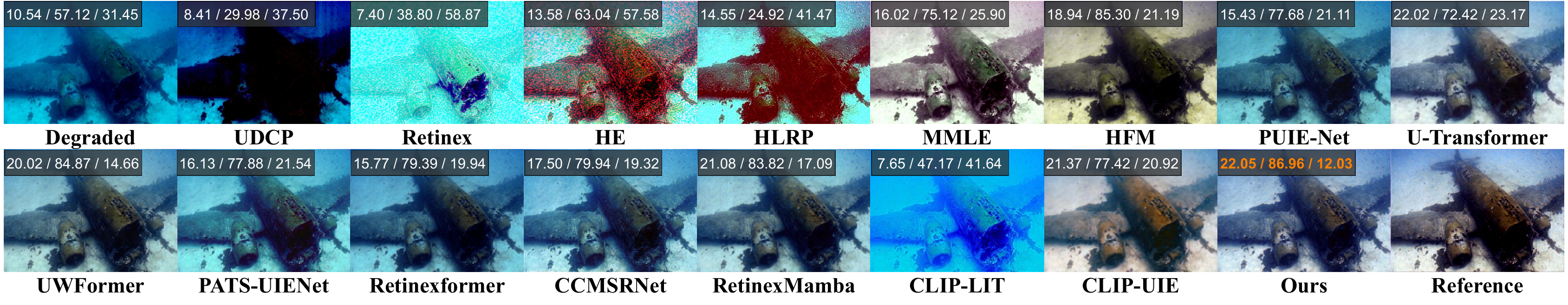}
 \caption{The results produced by 15 baselines and our method in terms of a degraded image in the Test-S110 test set. Here, the PSNR, SSIM and LPIPS values, computed between a degraded or enhanced image and the reference image, are shown at the top-left corner of the image.}
\label{fig:results_s110}
\end{figure*}

\begin{figure*}[!t]
\centering    \includegraphics[width=0.95\textwidth]{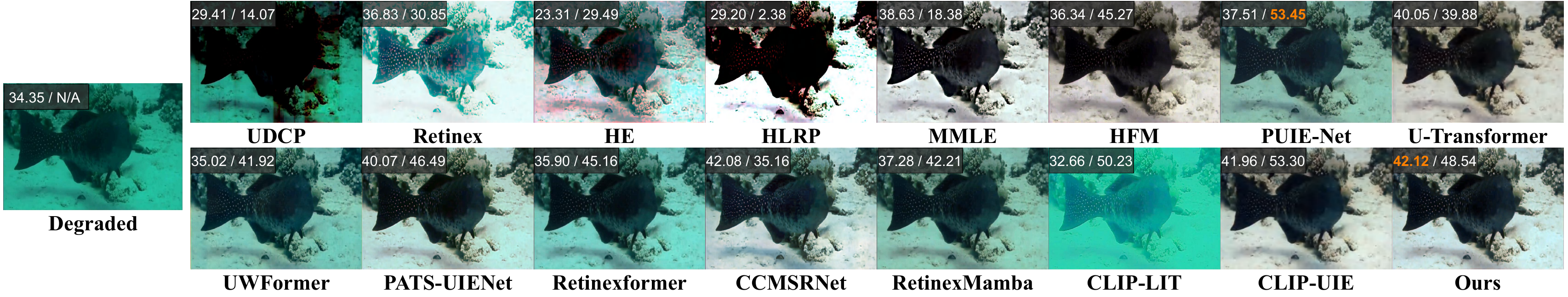}
 \caption{The results produced by 15 baselines and our method in terms of a degraded image in the Test-C60 test set. Here, both the PAUQA and UIF values are shown at the top-left corner of the degraded image and enhanced images.}
\label{fig:results_c60}
\end{figure*}

\begin{figure*}[!t]
\centering    \includegraphics[width=0.95\textwidth]{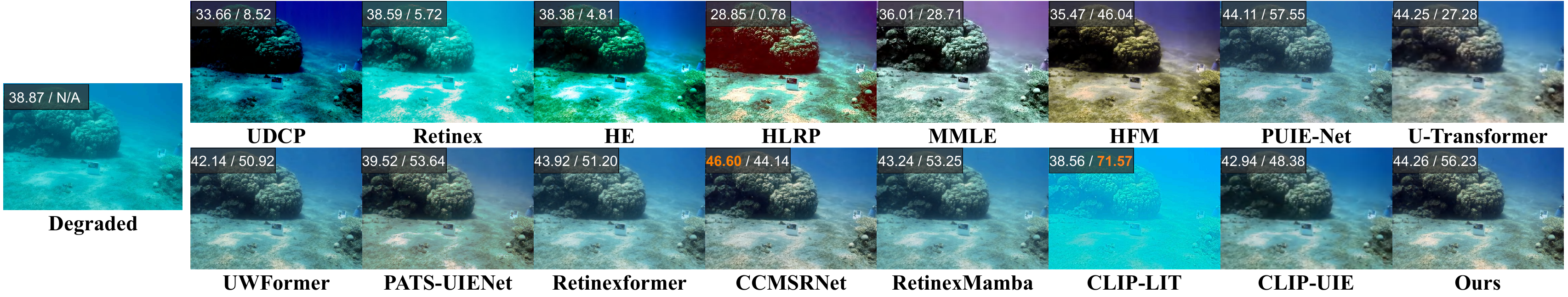}
\caption{The results produced by 15 baselines and our method in terms of a degraded image in the Test-R53 test set. Here, both the PAUQA and UIF values are shown at the top-left corner of the degraded image and enhanced images.}
\label{fig:results_r53}
\end{figure*}

The enhanced images produced by our method and 15 baselines on five test sets are shown in Figs. \ref{fig:results_l622} to \ref{fig:results_r53}, respectively. As can be observed, on the full-reference data sets, including Test-T622, Test-U80 and Test-S110, the proposed PSG-UIENet effectively improved different types of degraded images and produced results with natural and vivid colors. In contrast, traditional methods generally failed to achieve satisfactory performance. For instance, color reproduction of some traditional methods \cite{retinex, he, hlrp} was often unrealistic even though they generated images with high brightness. 

When the non-reference data sets, including Test-C60 and Test-R53, were used, the images enhanced by our method consistently exhibited superior visual quality, even in cases where the PAUQA \cite{pauqa} or UIF \cite{uif} scores were slightly lower than those produced by certain baselines \cite{puie, ccmsrnet, clip-lit}.
Although some baselines achieved higher scores in the LPIPS \cite{lpips}, PAUQA \cite{pauqa} and UIF \cite{uif} metrics, their visual results were still unsatisfactory. For example, CLIP-LIT \cite{clip-lit} produced the highest UIF \cite{uif} score across all five test sets; however, the enhanced images that it produced did not consistently show satisfactory visual quality, as illustrated in Figs. \ref{fig:results_l622} to \ref{fig:results_r53}.

\subsection{Ablation Study}
To evaluate the impact of different modules of the proposed method, we conducted three ablation experiments. For simplicity, only the Test-L622 was utilized. The evaluation was performed using full-reference metrics, including PSNR, SSIM and LPIPS, and a non-reference metric, i.e., PAUQA. 


\subsubsection{Effect of the Components of PSG-UIENet}
To investigate the impact of each core component in PSG-UIENet, we conducted an ablation experiment by individually removing the Illumination Estimator (w/o IE), Image Restorer (w/o IR), Text Aligner (w/o TA) and Cross-Attention FiLM Module (w/o CFM). As shown in Table~\ref{tab:ablation_components}, the removal of a component normally results in a performance decline, confirming the usefulness of each component. Specifically, the greatest drop can be observed in the PSNR and SSIM metrics when the IR has been removed, indicating its essential role in structural reconstruction. The removal of the TA achieved the lowest LPIPS value, but the PAUQA and PSNR values slightly reduced, suggesting that it enhances perceptual details at the cost of losing semantic or structural characteristics. Besides, the PAUQA and LPIPS values slightly decreased when the CFM was removed, underscoring its importance in semantic modulation. In contrast, the full PSG-UIENet configuration achieved the best balance across different metrics.

\begin{table}[t]
\begin{center}
\caption{Comparison between our method and its four variants obtained by removing a key component in PSG-UIENet.}
\label{tab:ablation_components}
\setlength{\tabcolsep}{2.3mm}{
\begin{tabular}{c|ccccc}
\toprule
\multirow{2}{*}{Method} & \multicolumn{5}{c}{Test-L622} \\
\cmidrule{2-6} 
& PSNR$\uparrow$   & SSIM$\uparrow$   & LPIPS$\downarrow$    & PAUQA$\uparrow$  & UIF$\uparrow$\\
\midrule
w/o IE      & 23.49 & 89.70 & 9.56 & \textcolor{blue}{\textit{51.27}} & 67.43 \\  
w/o IR      & 22.44 & 86.47 & 11.60 & 48.73 & \textcolor{red}{\textbf{77.70}} \\
w/o TA        & \textcolor{blue}{\textit{23.69}} & \textcolor{blue}{\textit{90.11}} & \textcolor{red}{\textbf{8.90}}  & 51.10 & 67.14 \\
w/o CFM       & 23.62 & 89.73 & 9.21 & 50.78 & 67.67 \\  
\midrule
Ours       & \textcolor{red}{\textbf{24.07}} & \textcolor{red}{\textbf{90.19}} & \textcolor{blue}{\textit{9.11}} & \textcolor{red}{\textbf{51.34}} & \textcolor{blue}{\textit{67.70}} \\ 
\bottomrule
\end{tabular}}
\end{center}
\end{table}

\subsubsection{Effect of the Guidance of Textual Descriptions} 
To evaluate the impact of the guidance of textual descriptions, we compared the full model with its two variants, in which one variant was derived by removing the text path (w/o Text) and the other variant was obtained by replacing the proposed cross-attention mechanism with the standard Multi-head Self-attention (MHSA) mechanism (w/ MHSA). As shown in Table~\ref{tab:ablation_text}, removing the text path led to a noticeable performance drop in PSNR, SSIM and LPIPS, despite a slightly higher PAUQA value being produced. This finding suggests that structural and semantic fidelity were compromised without textual descriptions even though the perceptual quality remained. In addition, the replacement of the cross-attention mechanism with the MHSA mechanism resulted in performance degradation across all metrics, further validating the effectiveness of our image-text fusion mechanism. These results demonstrate that the joint use of textual features is useful for improving UIE performance.

\begin{table}[t]
\begin{center}
\caption{Comparison between our method and its two variants obtained by removing the text path or replacing the cross-attention mechanism with the MHSA mechanism.}
\label{tab:ablation_text}
\setlength{\tabcolsep}{2.5mm}{
\begin{tabular}{c|ccccc}
\toprule
\multirow{2}{*}{Method} & \multicolumn{5}{c}{Test-L622} \\
\cmidrule{2-6} 
& PSNR$\uparrow$   & SSIM$\uparrow$   & LPIPS$\downarrow$    & PAUQA$\uparrow$  & UIF$\uparrow$\\
\midrule
w/o Text      & \textcolor{blue}{\textit{23.66}} & \textcolor{blue}{\textit{89.93}} & \textcolor{blue}{\textit{9.21}} & \textcolor{red}{\textbf{51.48}} & 67.41 \\  
w/ MHSA        & 23.25 & 89.52 & 9.61 & 51.03 & \textcolor{red}{\textbf{70.03}} \\
\midrule
Ours       & \textcolor{red}{\textbf{24.07}} & \textcolor{red}{\textbf{90.19}} & \textcolor{red}{\textbf{9.11}} & \textcolor{blue}{\textit{51.34}} & \textcolor{blue}{\textit{67.70}} \\ 
\bottomrule
\end{tabular}}
\end{center}
\end{table}

\subsubsection{Effect of the Masking Ratio} 
We evaluated the effect of five masking ratio ($\theta$) values, including 0, 0.25, 0.5, 0.75 and 1.0. As shown in Table~\ref{tab:ablation_mask}, the best performance was achieved when $\theta$ was set to 0.5 (default), indicating that a suitable masking ratio value effectively balances semantic learning and structural preservation. When masking was not applied ($\theta$ = 0), the performance of our method decreased slightly, particularly in terms of the PSNR and PAUQA metrics, suggesting the importance of forcing semantic completion. In contrast, overly aggressive masking ($\theta$ = 1) caused performance degradation across all metrics due to the lack of sufficient visual context. These results suggest that moderate masking encourages semantic-text interaction without compromising visual fidelity.

\begin{table}[t]
\begin{center}
\caption{Comparison of five masking ratio ($\theta$) values.}
\label{tab:ablation_mask}
\setlength{\tabcolsep}{2.3mm}{
\begin{tabular}{c|ccccc}
\toprule
\multirow{2}{*}{$\theta$} & \multicolumn{5}{c}{Test-L622} \\
\cmidrule{2-6} 
& PSNR$\uparrow$   & SSIM$\uparrow$   & LPIPS$\downarrow$    & PAUQA$\uparrow$  & UIF$\uparrow$\\
\midrule
0     & \textcolor{blue}{\textit{23.82}} & \textcolor{blue}{\textit{89.93}} & \textcolor{blue}{\textit{9.19}} & \textcolor{blue}{\textit{51.33}} & 67.23 \\  
0.25      & 22.08 & 88.78 & 9.73 & 50.34 & 67.55 \\
0.5 (Default)       & \textcolor{red}{\textbf{24.07}} & \textcolor{red}{\textbf{90.19}} & \textcolor{red}{\textbf{9.11}} & \textcolor{red}{\textbf{51.34}} & \textcolor{blue}{\textit{67.70}} \\
0.75       & 23.23 & 89.58 & 9.55 & 51.05 & \textcolor{red}{\textbf{68.52}} \\ 
1.0       & 22.87 & 88.99 & 10.29 & 50.32 & 67.52 \\  
\bottomrule
\end{tabular}}
\end{center}
\end{table}

\section{Conclusion}
\label{conclusion}
We proposed PSG-UIENet, a Physics-Semantics-Guided Underwater Image Enhancement Network, which combines the Retinex-based physical prior with the high-level semantic cues encoded in textual descriptions. This network integrates a prior-free illumination estimator and a dual-branch U-shaped restorer with cross-attention-based fusion, enabling effective multi-modal interaction and robust adaptation across diverse underwater scenarios. To support multi-modal learning, we constructed the first large-scale image-text underwater data set, namely, LUIQD-TD, which contains degraded-reference image pairs and associated scene-level textual annotations. We further introduced an Image-Text Semantic Similarity (ITSS) loss function for the purpose of promoting semantic consistency between enhanced images and textual descriptions. Extensive experiments on five test sets demonstrated that PSG-UIENet achieved superior or competitive results compared to 15 traditional and state-of-the-art approaches, validating the effectiveness of integrating physics-inspired modeling with language-based supervision.




\bibliographystyle{IEEEtran}
\bibliography{ref}


\begin{IEEEbiography}[{\includegraphics[width=1in,height=1.25in,clip,keepaspectratio]{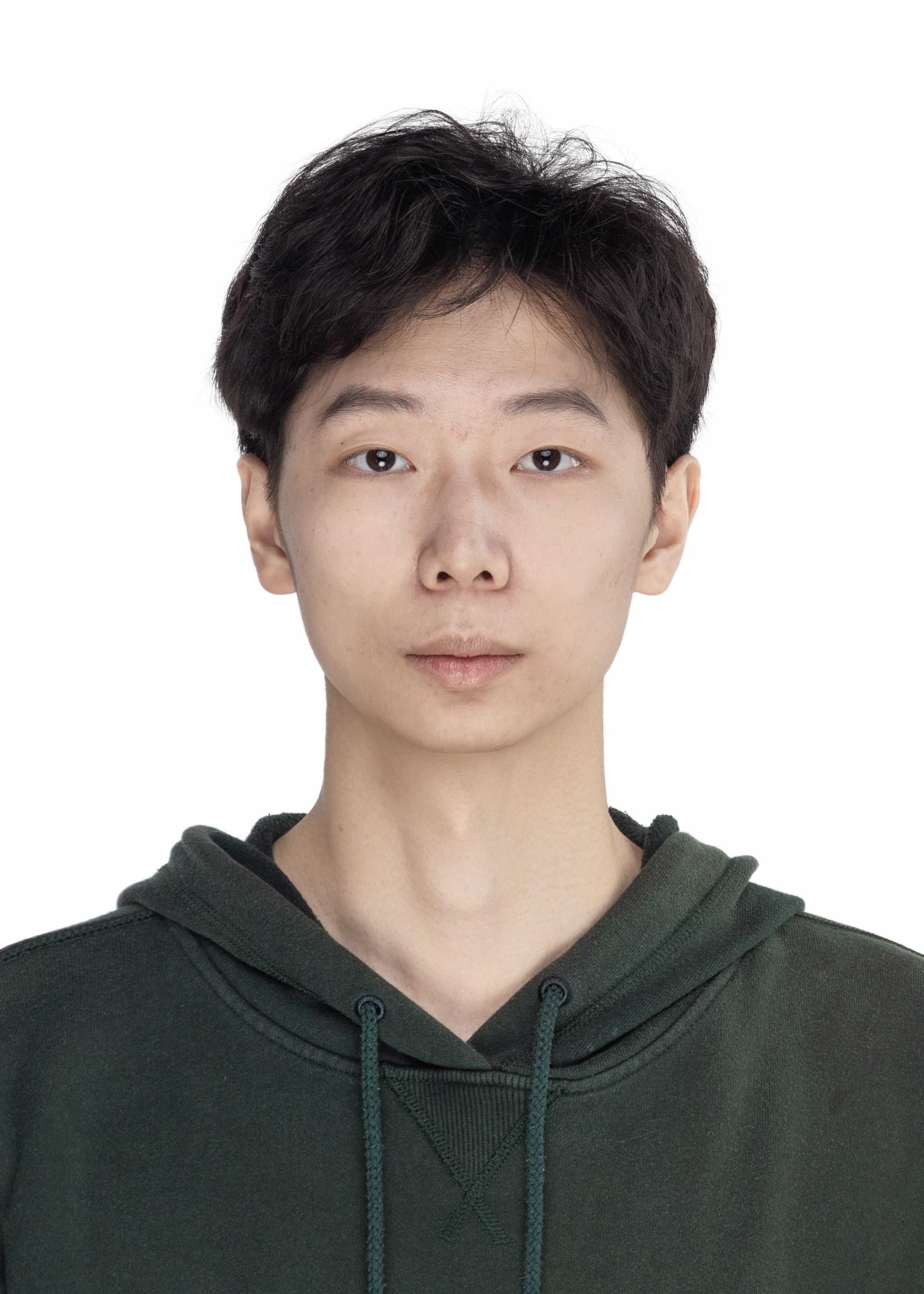}}]{Shixuan Xu} received the bachelor’s degree in Engineering from Lanzhou University of Finance and Economics (LZUFE), Lanzhou, Gansu, China, in 2022. He is currently pursuing the master’s degree in Artificial Intelligence at Ocean University of China. His research interests include computer vision, deep learning and image enhancement.
\end{IEEEbiography}


\begin{IEEEbiography}[{\includegraphics[width=1in,height=1.25in,clip,keepaspectratio]{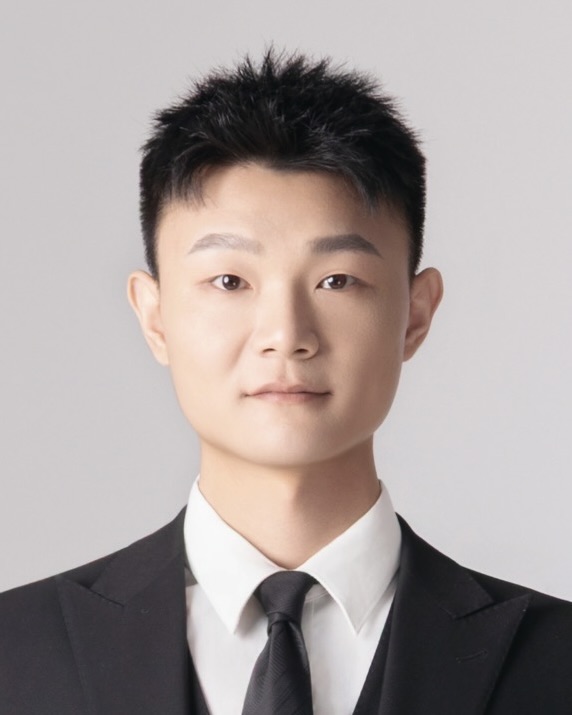}}]{Yabo Liu} received the Ph.D. degree in computer technology from Harbin Institute of Technology, Shenzhen, China, in 2025. From 2021 to 2025, he was a jointly supervised Ph.D. candidate by Harbin Institute of Technology and Peng Cheng Laboratory. He is currently a lecturer with the School of Artificial Intelligence, Ocean University of China, Qingdao, China. His current research interests include computer vision, transfer learning, machine learning, and multi-modal learning. 
\end{IEEEbiography}


\begin{IEEEbiography}[{\includegraphics[width=1in,height=1.25in,clip,keepaspectratio]{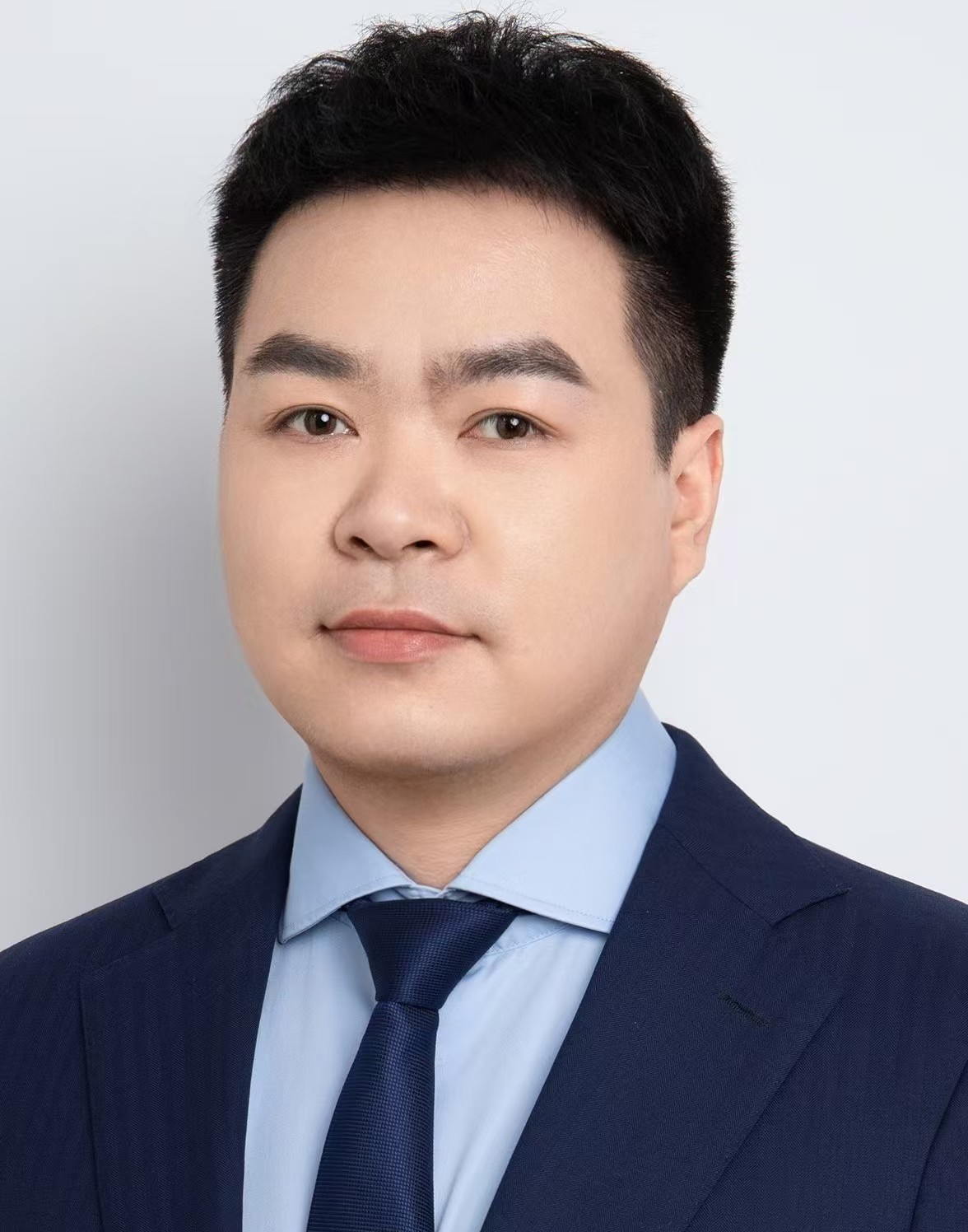}}]{Chao Huang} received the Ph.D. degree in computer science and technology from Harbin Institute of Technology, Shenzhen, China, in 2022. From 2019 to 2022, he was a visiting scholar with Peng Cheng Laboratory, Shenzhen. He is currently an Assistant Professor with the School of Cyber Science and Technology, Sun Yat-sen University, Shenzhen. So far, he has published over 60 technical papers in prestigious international journals and conferences. His research interests include anomaly detection, multimedia analysis, object detection, image/video compression, and deep learning. Dr. Huang received the Distinguished Paper Award of AAAI 2023, and his dissertation was nominated for Harbin Institute of Technology’s Outstanding Dissertation Award. He serves as an Associate Editor for Pattern Recognition and serves/served as the reviewer/ PC member for several top-tier journals and conferences, including IEEE TPAMI, TIP, TIFS, TNNLS, ACM CSUR, CVPR, ICCV, ECCV, ICML, NeurIPs, ICLR, AAAI, IJCAI, and ACM Multimedia.
\end{IEEEbiography}


\begin{IEEEbiography}[{\includegraphics[width=1in,height=1.25in,clip,keepaspectratio]{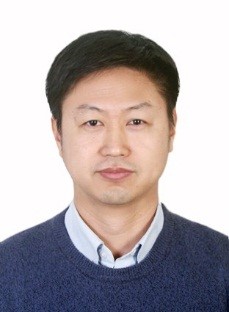}}]{Junyu Dong} received the B.Sc. and M.Sc. degrees from the Department of Applied Mathematics, Ocean University of China, Qingdao, China, in 1993 and 1999, respectively, and the Ph.D. degree in image processing from the Department of Computer Science, Heriot-Watt University, U.K., in 2003. He joined Ocean University of China in 2004. He is currently a Professor and the Dean of the Faculty of Information Science and Engineering, Ocean University of China. His research interests include computer vision, underwater image processing, and machine learning.
\end{IEEEbiography}


\begin{IEEEbiography}[{\includegraphics[width=1in,height=1.25in,clip,keepaspectratio]{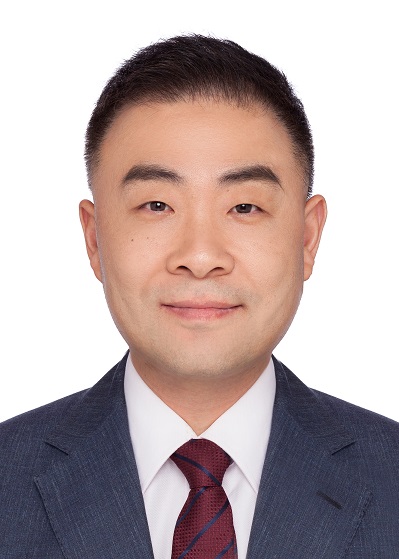}}]{Xinghui Dong} received the PhD degree from Heriot-Watt University, U.K., in 2014. He worked with the Centre for Imaging Sciences, the University of Manchester, U.K., between 2015 and 2021. Then he jointed Ocean University of China in 2021. He is currently a professor at the Ocean University of China. His research interests include computer vision, defect detection, texture analysis, underwater image processing and visual perception.
\end{IEEEbiography}

\end{document}